\documentclass[letterpaper, 10 pt, conference]{ieeeconf}

 \usepackage{ownstyle3}
 \usepackage[colorlinks,citecolor=blue,urlcolor=blue]{hyperref}
 
 \usepackage{flushend}
 
 \newcommand{\kmax}{k_\text {max}}

\newcommand{\overbar}[1]{\mkern 1.5mu\overline{\mkern-1.5mu#1\mkern-1.5mu}\mkern 1.5mu}

 \title{Learning graphical models from \\ the Glauber dynamics}

\author{
Guy Bresler\textsuperscript{1}  \qquad David Gamarnik\textsuperscript{2} \qquad Devavrat Shah\textsuperscript{1}
\\ \\
Laboratory for Information and Decision Systems
\\
Department of Electrical Engineering and Computer Science\textsuperscript{1} \\
Operations Research Center and Sloan School of Management\textsuperscript{2}
\\ 
Massachusetts Institute of Technology\\
\texttt{\{gbresler,gamarnik,devavrat\}@mit.edu}
}

\begin{document}
\maketitle
\begin{abstract}
In this paper we consider the problem of learning undirected graphical models from data generated according to the Glauber dynamics. The Glauber dynamics is a Markov chain that sequentially updates individual nodes (variables) in a graphical model and it is frequently used to sample from the stationary distribution (to which it converges given sufficient time).   Additionally, the Glauber dynamics is a natural dynamical model in a variety of settings. This work 
deviates from the standard formulation of graphical model learning in the literature, where one assumes access to i.i.d. samples from the distribution. 

Much of the research on graphical model learning has been directed towards finding algorithms with low computational cost.
As the main result of this work, we establish that the problem of reconstructing binary pairwise graphical models is \emph{computationally tractable} 
when we observe the Glauber dynamics. Specifically, we show that a binary pairwise graphical model on $p$ nodes with maximum degree $d$ can be learned in time 
$f(d)p^2\log p$, for a function $f(d)$, using nearly the information-theoretic minimum number of samples. 
\end{abstract}

\section{Introduction}

Examples of data one might usefully model as being generated according to a Markov process include the dynamics of agents in a coordination game, the fluctuations of stocks or other financial data, behavior of users in a social network, and spike patterns in neural networks. 

The focus of this paper is on learning the nature of Markovian dynamics from observed data governed by local interactions. Concretely, we suppose that such local interactions are represented by a 
graphical model. We observe a single-site dynamics, specifically the so-called Glauber dynamics, and wish to learn the graph underlying the model. 

This work fits within a broader theme of learning graphical models from data, a problem traditionally posed assuming access to i.i.d. samples from the model. While the assumption of i.i.d. samples 
makes sense as an abstraction (as well as in some practical scenarios), observations in many settings are correlated over time and in this case it is more natural to assume that samples are generated according to a Markov process. In general the distribution of such samples can be far from i.i.d.

The problems of learning and of generating samples are known to be related. 
On one hand, learning graphical models from i.i.d. samples is algorithmically challenging \cite{BGS14b,Montanari2014,bento2009graphical}, and on the other hand, generating samples from distributions represented by graphical models
is hard in general \cite{sly2012computational}. In the literature, much work has focused on trying to find low-complexity algorithms, both for learning as well as for generating samples, under various restrictions to the graphical model. Interestingly, related conditions (such as spatial and temporal mixing) have turned out to be central to most approaches.

Learning graphical models from i.i.d. samples appears to be challenging when there are correlations between variables on a global scale, as this seems to require a global procedure. Our results show that observing a \emph{local} process allows to learn distributions with global correlations by \emph{temporally} isolating the local structure.

\subsection{Complexity of graphical model learning}
A number of papers, including \cite{abbeel2006learning}, \cite{BMS08}, and~\cite{csiszar2006consistent}
have suggested finding each node's neighborhood by exhaustively searching over candidate neighborhoods and checking conditional independence.  
For graphical models on $p$ nodes of maximum degree $d$, such a search takes time (at least) on the order of $p^d$. As $d$ grows, the computational cost becomes prohibitive, and much effort by the community has focused on trying to find algorithms with lower complexity. 

Writing algorithm runtime in the form $f(d) p^{c(d)}$, for high-dimensional (large $p$) models the exponent $c(d)$ is of primary importance, and we will think of low-complexity algorithms as having an exponent $c(d)$ that is bounded by a constant independent of $d$.

Previous works proposing low-complexity algorithms either restrict the graph structure or the nature of the interactions between variables. The seminal paper of Chow and Liu \cite{chow1968approximating} makes a model restriction of the first type, assuming that the graph is a tree; generalizations include to polytrees~\cite{dasgupta1999learning}, hypertrees~\cite{srebro2001maximum}, tree mixtures~\cite{anandkumar2012learningB},
and others. Among the many possible assumptions of the second type, the correlation decay property (CDP) is distinguished: 
nearly all existing low-complexity algorithms require the CDP~\cite{bento2009graphical}. An exception is~\cite{BGS14a}, which shows a family of antiferromagnetic models that can be learned with low complexity despite strongly violating the CDP.

Informally, a graphical model is said to have the correlation decay property (CDP) if any two variables $\sig_s$ and $\sig_t$ are asymptotically independent as the graph distance between $s$ and $t$ increases.
The CDP is known to hold for a number of pairwise graphical models in the so-called high-temperature regime, including Ising, hard-core lattice gas, Potts (multinomial), and others (see the survey article \cite{gamarnikcorrelation} as well as, e.g., 
 \cite{dobrushin1970prescribing,dobrushin1985constructive,salas1997absence,gamarnik2013correlation,gamarnik2007correlation,bandyopadhyay2008counting,Weitz}). 

It was first observed in \cite{BMS08} that it is possible to efficiently learn models with (exponential) decay of correlations, under the additional assumption that neighboring variables have correlation bounded away from zero. A variety of other papers including \cite{netrapalli2010greedy,raygreedy,anandkumar2012high,wu_2013_learning} give alternative low-complexity algorithms, but also require the CDP. 
A number of structure learning algorithms are based on convex optimization, such as Ravikumar et al.'s \cite{ravikumar2010high} approach using regularized node-wise logistic regression. While this algorithm is shown to work under certain incoherence conditions and does not explicitly require the CDP, Bento and Montanari \cite{bento2009graphical} showed through a careful analysis that the algorithm provably fails to learn ferromagnetic Ising models on simple families of graphs without the CDP. Other convex optimization-based algorithms such as \cite{lee2006efficient,jalali2011learning,jalali2011learning2}
require similar incoherence or restricted isometry-type conditions that are difficult to interpret in terms of model parameters, and likely also require the CDP.  

Most computationally efficient sampling algorithms (which happen to be based on the Markov Chain Monte Carlo method) require a notion of temporal mixing and this is closely related to spatial mixing or a version of the CDP (see, e.g., \cite{stroock1992logarithmic,dyer2004mixing,martinelli1994approach}). Thus, under a class of ``mixing conditions", we can both generate (i.i.d.) samples efficiently as well as learn graphical models efficiently from such i.i.d. samples. 

\subsection{Main results}

We give an algorithm that learns the graph structure underlying an arbitrary undirected binary pairwise graphical model from the Glauber dynamics, even without any mixing or correlation decay property. Concretely, in Theorem~\ref{t:Structure} we show that the algorithm learns the graph underlying any undirected binary pairwise graphical model over $p$ nodes with maximum vertex degree $d$, given $\Omega(\log p)$ updates of the Glauber dynamics per node,
starting from any initial state, with runtime $f(d) p^2 \log p$. The number of samples required by the algorithm is nearly information-theoretically optimal, as shown in the lower bound of Theorem~\ref{t:lower}. 

\subsection{Other related work}

Several works have studied the problem of learning the graph underlying a random process for various processes. These include learning from epidemic cascades \cite{netrapalli2012learning,rodriguez2011uncovering,myers2010convexity} and learning from delay measurements \cite{anandkumar2011topology}. 
Another line of research asks to find the source of infection of an epidemic by observing the current state, where the graph is known \cite{shah2011rumors,prakash2014efficiently}. 

More broadly, a number of papers in the learning theory community have considered learning functions (or concepts) from examples generated by Markov chains, including \cite{aldous1990markovian,bartlett1994exploiting,bshouty2005learning,gamarnik2003extension}. The present paper is similar in spirit to that of Bshouty et al. \cite{bshouty2005learning} showing that it is relatively easy to learn DNF formulas from examples generated according to a random walk as compared to i.i.d. samples. 

The literature on the Glauber dynamics is enormous and we do not attempt to summarize it here. However, we remark that the Glauber dynamics is equivalent to a model of noisy coordination games and has been studied in that context by various authors: Saberi and Montanari \cite{montanari2010spread} studied the impact of graph structure on rate of adoption of innovations,
Berry and Subramanian~\cite{berry2012spotting} studied the problem of inferring the early adopters from an observation at a later time.

\subsection{Outline}

The rest of the paper is organized as follows. In Section~\ref{s:Setup} we define the model and formulate the learning problem. 
In Section~\ref{s:alg} we present our structure learning algorithm and analysis.
Then in Section~\ref{s:lower} we give an information-theoretic lower bound on the number of samples necessary in order to reconstruct with high probability.

\section{Problem statement}\label{s:Setup}

\subsection{Ising model.}

We consider the Ising model on a graph $G=(V,E)$ with $|V|=p$. The notation $\partial i$ is used to denote the set of neighbors of node $i$, and the degree $|\partial i|$ of each node $i$ is assumed to be bounded by $d$. 
To each node $i\in V$ is associated a binary random variable (spin) $\sigma_i$. Each configuration of spins $\sig\in \{-1,+1\}^V$ is assigned probability according to the Gibbs distribution
\begin{equation}\label{e:Ising}
\P(\sig) = \frac1Z \exp\bigg(\sum_{\{i,j\}\in E} \theta_{ij} \sig_i\sig_j\bigg)\,.
\end{equation}
Here $Z$ is the partition function and serves to normalize the distribution. The distribution is parameterized by the vector of edge couplings $(\theta_{ij})\in \R^E$, assumed to satisfy 
$$
\alpha \leq |\theta_{ij} | \leq \beta\quad \text {for } \{i,j\}\in E
$$
for some constants $0<\al\leq \beta$. We can alternatively think of $\theta\in \R^{p\choose 2}$, with $\theta_{ij}=0$ if $\{i,j\}\notin E$. For a graph $G$, let  
\begin{align*}
\Omega_{\alpha,\beta}(G) = \{\theta\in \R ^{p\choose 2} &: \alpha\leq |\theta_{ij}|\leq \beta\text { if } \{i,j\}\in E, \\& \qquad\qquad \text {and }\theta_{ij} = 0  \text{ otherwise}\}
\end{align*}
be the set of parameter vectors corresponding to~$G$. 

The model~\eqref{e:Ising} does not have node-wise parameters (that is, the external field is zero); while we restrict to this case for simplicity, similar results to those presented hold with suitable minor modifications to accommodate nonzero external fields.

The distribution specified in $\eqref{e:Ising}$ is a \emph{Markov random field}, and an implication is that each node is conditionally independent of all other nodes given the values of its neighbors. This allows to define a natural Markov chain known as the Glauber dynamics. 

\subsection{The Glauber dynamics.}\label{ss:Glauber}
 The Glauber dynamics (also sometimes called the Gibbs Sampler) is a natural and well-studied reversible Markov chain defined for any Markov random field. For mathematical convenience we use both the continuous-time and discrete-time versions. We describe here the continuous-time dynamics, writing $\sig^t$ for the configuration at time $t\geq 0$. The process is started at some arbitrary (possibly random) initial configuration $\sig^0\in \{-1,+1\}^p$, and each node is updated at times given by an independent Poisson process of rate one. If spin $\sig_i$ is updated at time $t$, it takes on value $+1$ with probability
\begin{align}
\P(\sig_i=+1\,|\sig^{t}_{V\setminus \{i\}}) = 
 \frac{\exp\big(2 \sum_{j\in \partial i} \theta_{ij} \sigma_j^t \big)}{1+\exp\big(2 \sum_{j\in \partial i} \theta_{ij} \sigma_j^t \big)}
\,,\label{e:Glauber}
\end{align}
and is $-1$ otherwise. Notably, each spin update depends only on neighboring spins. 
Equation~\eqref{e:Glauber} and the bounded coupling assumption $|\theta_{ij}|\leq \beta$ implies that for any $x\in \{-1,+1\}^{\partial i}$,
\begin{align}
\label{e:minProb}
\min\{\P(\sig_i =+1 \,| \sig^t_{\partial i} = x), \P(\sig_i =-1 \,| \sig^t_{\partial i}=x)\} 
\geq \tfrac12 e^{-2\beta d}\,.
\end{align}
This is a lower bound on the randomness in each spin update and will be used later. 

The Glauber dynamics can be simulated efficiently for any bounded-degree undirected graphical model, and it is a plausible generating process for observed samples in various settings. One can check that the Gibbs distribution \eqref{e:Ising} is stationary for the Glauber dynamics. If the dynamics quickly approaches stationarity (that is, the {mixing time} is small), then it can be used to simulate i.i.d. samples from~\eqref{e:Ising}. But there are families of graphs for which any local Markov chain, including the Glauber dynamics, is known to converge exponentially slowly (see, e.g., \cite{sly2012computational}), and moreover the availability of i.i.d. samples violates conjectures in complexity theory in that it allows to approximate the partition function \cite{jerrum1986random}. While it is difficult to imagine nature producing i.i.d. samples from such models, there is no such issue with the Glauber dynamics (or any other local Markov chain).

\subsection{Graphical model learning}

Our goal is to learn the graph $G=(V,E)$ 
underlying a graphical model of the form~\eqref{e:Ising}, given access to observations from the Glauber dynamics. We assume that the identity of nodes being updated is known; learning without this data is potentially much more challenging, because in that case information is obtained only when a spin flips sign, which may occur only in a small fraction of the updates. 

For the purposes of recording the node update sequence it is more convenient to work with a discrete time (heat-bath) version of the chain, where each sample is taken immediately after a node is updated. In this case we denote the sequence of $n$ samples by $\sigma^{(1)},\sigma^{(2)},\dots, \sigma^{(n)}$ and the corresponding node identities at which updates occur by $I^{(1)},I^{(2)},\dots, I^{(n)}$. The value of $I^{(1)}$ is arbitrarily set to (say) one since the first configuration does not arise from a node update. The sequence of $n$ samples is denoted by 
\begin{equation}\label{e:X}
X=(\sigma^{(l)}, I^{(l)})_{1\leq l\leq n}
\end{equation} 
and is therefore an element of the product space $$\XX=(\{-1,+1\}^p)^n\times [p]^{n}\,.$$
We suppose that the continuous-time chain is observed for $T$ units of time, 
so there are in expectation $Tp$ spin updates. This number is tightly concentrated around the mean, and 
our arguments are not sensitive to a small amount of randomness in the number of samples $n$, so for convenience we deterministically set $n=Tp$.

\newcommand{\Gpd}{{\GG_{p,d}}}

As mentioned before, the underlying graph $G$ is assumed to have maximum node degree bounded by $d$, and we denote the set of all such graphs on $p$ nodes by $\Gpd$. A \emph{structure learning algorithm} is a (possibly randomized) map
$$
\phi:(\{-1,+1\}^p)^n\times [p]^{n}\to \Gpd\,.
$$
The performance of a structure learning algorithm is measured using the zero-one loss, and the risk under some vector $\theta\in \Omega(G)$ of parameters corresponding to a graph $G\in \Gpd$ is given by 
$$
\P_{\theta} (\phi(X)\neq G)\,.
$$
The minimax risk is the best algorithm's worst-case risk (probability of error) over graphs and corresponding parameter vectors, namely
$$
R_{p,d,n}\triangleq\min_\phi\max_{G\in \Gpd\atop \theta\in \Omega(G)} \P_{\theta} (\phi(X)\neq G)\,.
$$ 

The basic questions we seek to address are what triples $n,p,d$ result in the minimax risk $R_{p,d,n}$ tending to zero as these parameters tend to infinity, and can we find an efficient algorithm.

\section{Structure learning algorithm}
\label{s:alg}
\subsection{Idealized test}

\newcommand{\pp}{{p^+}}
\newcommand{\pmi}{{p^-}}
We determine the presence of edges in a decoupled manner, focusing on a single pair of nodes $i$ and $j$. 
Our test is based on the identity (derived via Eq.~\eqref{e:Glauber})
\begin{equation}\label{e:BasicIdentity}
e^{4\theta_{ij}} = \frac{\pp(1-\pmi)}{\pmi (1-\pp)}\,,
\end{equation}
where for an arbitrary assignment $x_{\partial i\setminus\{j\}}$ we define
\begin{align*}
\pp(x_{\partial i\setminus\{j\}}) &=\P(\sig_i = +1 | \sig_{\partial i \setminus\{j\}}=x_{\partial i\setminus\{j\}}, \sig_j=+1) \\
 \pmi(x_{\partial i\setminus\{j\}})&=\P(\sig_i = +1 | \sig_{\partial i \setminus\{j\}}=x_{\partial i\setminus\{j\}}, \sig_j=-1)
\end{align*}
(We will often leave implicit the dependence of $\pp$ and $\pmi$ on $x_{\partial i\setminus \{j\}}$.)
The identity~\eqref{e:BasicIdentity} holds whether or not the edge $\{i,j\}$ is present, since $\{i,j\}\notin E$ implies $\theta_{ij}=0$, and this agrees with $\sig_i$ and $\sig_j$ being conditionally independent given $\sig_{\partial i}$ (in which case $\pp=\pmi$). 

Instead of attempting to estimate the right-hand side of \eqref{e:BasicIdentity} from samples, we claim that if our goal is merely to decide between $\theta_{ij}=0$ and $|\theta_{ij}|\geq \alpha$, it suffices to estimate the much simpler quantity $\pp - \pmi$. This will be justified using the following bound.

\begin{lemma}\label{l:smallIneq}
Let $a$ and $b$ be real numbers with $0<a\leq b<1$ and $a\leq \frac12$. Then
$$
 b-a \leq \frac{b(1-a)}{a(1-b)} - 1\leq \frac{b-a}{a(1-b)^2}\,.
$$
\end{lemma}
\vspace{3mm}
\begin{proof}
Let $g(y) = \frac{y}{1-y}$. 
Then for $y\in (0,1)$,
$
g'(y) = (1-y)^{-2} >1
$. 
It follows that $g'(y)\leq (1-b)^{-2}$ for $y\in [a,b]$ and also from
$a\leq \frac12$ we get $a\inv \geq \frac{1-a}a \geq 1$. Combining these ingredients gives
\belowdisplayskip=-12pt
\begin{align*}
b-a&\leq g(b) - g(a) \leq \frac{1-a}a \left(\frac b{1-b} - \frac a{1-a}\right) \\&=
\frac{b(1-a)}{a(1-b)} - 1 
\leq \frac1a (g(b) - g(a))
\leq \frac{b-a}{a(1-b)^2}\,.
\end{align*}
\end{proof}
\medskip

Let us momentarily assume that $\pp\geq \pmi$ and $\pmi\leq \frac12$.
The conditional probability lower bound~\eqref{e:minProb} implies $\min\{1-\pp,\pmi\}\geq \frac12 e^{-2\beta d}$. Lemma~\ref{l:smallIneq} together with 
\eqref{e:BasicIdentity} gives
\begin{equation*}
\pp - \pmi\leq e^{4\theta_{ij}} - 1\leq 8e^{8\beta d} (\pp - \pmi)\,.
\end{equation*}
The assumption $\pmi\leq \frac12$ is without loss of generality by replacing $\pmi$ and $\pp$ by $1-\pmi$ and $1-\pp$, respectively. If $\pp<\pmi$, which happens if and only if $\theta_{ij}<0$, we get a similar sequence of inequalities:
\begin{equation*}
\pmi - \pp \leq e^{-4\theta_{ij}} - 1\leq 8e^{8\beta d} (\pmi - \pmi)\,.
\end{equation*}
These can be combined to give
\begin{equation}\label{e:squeeze}
\sign(\theta_{ij})(\pp - \pmi)\leq e^{4|\theta_{ij}|} - 1\leq \sign(\theta_{ij})8e^{8\beta d} (\pp - \pmi)\,.
\end{equation}
We emphasize that this inequality holds for any assignment $x_{\partial i\setminus \{j\}}$.

It turns out to be possible to crudely estimate the quantity $\pp - \pmi$ in~\eqref{e:squeeze} to determine if it is equal to zero. It is important that the sign of $\pp - \pmi$ does not depend on the configuration $x_{\partial i\setminus \{j\}}$, as this allows to accumulate contributions from many samples. The following scenario gives intuition for why sequential updates allow to do this. Suppose that $\sigma_i$ is updated, followed by $\sigma_j$ flipping sign, followed by yet another update of $\sigma_i$, with no other spins updated. Since only $\sigma_j$ has changed in between updates to $\sigma_i$, we can hope to get an estimate of the effect of $\sigma_j$ on $\sigma_i$. To produce the sequence of events just described requires observing the process for $\Omega(p^2)$ time; we next show how to achieve a similar outcome sufficient for learning the structure, in time only $O(\log p)$.

\newcommand{\sigt}[2]{\sigma^{[#1,#2)}}

\subsection{Estimating edges}
We define a few events to be used towards estimating the effect of an edge, as captured by $|\pp - \pmi|$.
To this end, consider restriction of the process $\sigma^t\in \{-1,+1\}^p$ to an interval, written as 
$$\sigt {t_1}{t_2} = (\sigma^t)_{t_1\leq t<t_2}\,.$$
For a positive number $L$, let $A_{ij}(\sigt0L)$ be the event that node $i$ is selected at least once in the first $L/3$ time-units but not node $j$, node $j$ is selected at least once in the second $L/3$ time-units but not node $i$, and node $i$ is selected at least once in the final $L/3$ time-units but not node $j$.
It is immediate from the Poisson update times that
\begin{equation}\label{e:A1prob}
\P(A_{ij}(\sigt 0L) )= [(1-e^{-L/3})e^{-L/3})]^3:=q\,.
\end{equation}
(We denote this quantity by $q$ since it will be used often.)
Next, define the event that $\sig_j$ is opposite at time $L/3$ versus $2L/3$,
$$B_{ij}(\sigt 0 L) = \{\sigma_j^{L/3}\neq \sigma_j^{2L/3}\}\,,$$
and take the intersection of the two events,
$$
C_{ij}(\sigt 0L) = A_{ij}(\sigt0L) \cap B_{ij}(\sigt0L)\,.
$$
Whenever $\sig_j$ is updated, by Equation~\eqref{e:minProb} both the probabilities of flipping or staying the same are at least $\frac12e^{-2 d\beta}$, regardless of the states of its neighbors. It follows that the last update of $\sigma_j$ in the interval $[\frac L3, \frac{2L}3]$ has at least probability $\frac12e^{-2 d\beta}$ of being opposite to $\sigma_j^{L/3}$, so 
\begin{equation}\label{e:AprobEst}
\P(C_{ij}) = \P(A_{ij})\cdot \P(B_{ij}|A_{ij}) \geq 
\tfrac12\P(A_{ij})e^{-2 d\beta}\,.
\end{equation}
Note that determining the occurrence of $C_{ij}$ does not require knowing anything about the graph.

We now define the statistic that will be used to estimate presence of a given edge: For each $k\geq 1$ and $1\leq i< j\leq p$, let 
\begin{align*}
&X_{ij}^{(k)}=X_{ij}(\sigt{(k-1)L}{kL}) \\&= \ind_{C_{ij}(\sigt{(k-1)L}{kL})}(-1)^{\ind\{\sig_j^{L/3}=+1\}} (\sigma_i^{L/3} - \sigma_i^{L})\,.
\end{align*}
The value $X_{ij}^{(k)}\in \{-2,0,+2\}$ can be computed by an algorithm with access to the process $\sigt{(k-1)L}{kL}$.
The idea is that $\E X_{ij}^{(k)}$ gives a rough estimate of the effect of spin $j$ on spin $i$ by counting the number of times $\sig_i$ has differing updates when $\sig_j$ has changed. It is necessary that few or no neighbors of $i$ are updated during the time-interval, as these changes could overwhelm the effect due to $\sig_j$. We will see later that choosing $L$ sufficiently small ensures this is usually the case. 

\subsection{Structure learning algorithm}

We now present the structure learning algorithm. In order to determine presence of edge $\{i,j\}$ the algorithm divides up time into intervals of length $L$, estimates $\E X_{ij}$ from the intervals, and compares $|\E X_{ij}|$ to a threshold~$\tau$. 

\begin{algorithm}
\caption{\textsc{GlauberLearn}$(\sigt0T,L,\tau)$} 
\texttt{1: Let $\widehat E = \varnothing$ and $\kmax = \lfloor T/L\rfloor$.\\
2: \texttt{For} $1\leq i< j\leq p$
\\
3: \quad \texttt{If} $|\frac1{\kmax} \sum_{k = 1}^{\kmax} X_{ij}^{(k)}| \geq \tau$\\
4: \quad \texttt{Then} add edge $\{i,j\}$ to $\widehat E$. \\
5: \texttt{Output} $\widehat E$. }
\end{algorithm}

\begin{theorem}\label{t:Structure}
Consider the Ising model \eqref{e:Ising} on a graph $G$ with maximum degree $d$ and couplings bounded as $\alpha\leq |\theta_{ij}|\leq \beta$. Let $\sigt0T$ denote the continuous-time Glauber dynamics started from any configuration $\sigma^0$. 
If 
$$L= \frac{\alpha}{16d}e^{-10d\beta},\quad \tau = 3Ldq,\quad T\geq \frac{10^6 e^{20d\beta}}{\alpha^2} \log p\,,$$
where
$
q = \P(A_{ij})=[(1-e^{-L/3})e^{-L/3})]^3\,,
$
 then \textsc{GlauberLearn} outputs the correct edge set with probability $1-\frac1p$ with runtime $O(p^2\log p)$. 
\end{theorem}
\smallskip

In the remainder of this section we work towards proving Theorem~\ref{t:Structure}. 
We first bound the runtime of the algorithm. Suppose that when the samples are collected, they are stored as a list for each node giving the times the node is updated and the new value. Each computation in Line~3 takes time $O(\log p)$, and this is done for $O(p^2)$ pairs $i,j$, which gives the stated runtime.

Since the Glauber dynamics is time-homogeneous (and Markov), $\E(X_{ij}^{(k)}|\sig^{(k-1)L}=x)$ does not depend on the index $k$. 
Hence, we use the shorthand $\E_x X_{ij}$ for $\E(X_{ij}^{(k)}|\sig^{(k-1)L}=x)$ and similarly for $\P_x (\cdot ) = \P(\cdot |\sig^{(k-1)L}=x )$. 

Let $D_{ij}(\sigt {t_1} {t_2} )$ be the event that none of the neighbors of $i$, aside from possibly $j$, are selected in time-interval $[t_1,t_2)$.
Since $D_{ij}$ depends on disjoint Poisson clocks from those determining $A_{ij}$, the two events are independent (however, $D_{ij}$ is not necessarily independent of $B_{ij}$). 
It is immediate, again from the Poisson times of updates, that
\begin{align}
\P(D_{ij}(\sigt {(k-1)L}{kL}))&= \P(D_{ij}(\sigt 0L))\nonumber \\&=\big( e^{-L}\big)^{|\partial i\setminus\{j\}|} \geq \big( e^{-L}\big)^{d}\,.\label{e:Bprob}
\end{align}

At this point it is possible to make a connection to the idealized edge test formula \eqref{e:squeeze}. 
Conditioning on $D_{ij}$, our edge statistic has expectation
\begin{align}
&\E_x(X_{ij}|D_{ij})\nonumber \\&= \E_x(X_{ij}|C_{ij},D_{ij}) \cdot \P_x(C_{ij}|D_{ij})\nonumber
\\ &=\E_x \big( (-1)^{\ind\{\sig_j^{L/3}=+1\}} (\sigma_i^{L/3} - \sigma_i^{L})|C_{ij},D_{ij}
\big) \cdot \P_x(C_{ij}|D_{ij})\nonumber
\\&=
2  \big( \P(\sigma_i=+1 | \sigma_{\partial i\setminus \{j\}} = x_{\partial i\setminus \{j\}}, \sigma_j = +1)\nonumber
 \\&\quad - \P(\sigma_i=+1 | \sigma_{\partial i\setminus \{j\}} = x_{\partial i\setminus \{j\}} , \sigma_j = -1)
\big) \cdot \P_x(C_{ij}|D_{ij})\nonumber
\\&= 2\big(\pp(x_{\partial i\setminus \{j\}}) - \pmi(x_{\partial i\setminus \{j\}})\big)\P_x(C_{ij}|D_{ij})
 \,.\label{e:edgeEffectEst}
\end{align}
Of course, without knowing the neighbors of $i$ it is not clear whether or not event $D_{ij}$ has occurred, but as shown next in Lemma~\ref{l:goodEstimate}, if $L$ is small enough, then $D_{ij}$ occurs frequently and $X_{ij}$ gives a good estimate.

\begin{lemma}\label{l:goodEstimate}
We have the following estimates:
\begin{enumerate}
\item[(i)] If $\{i,j\}\in E$, then for any $x\in \{-1,+1\}^p$,
\begin{align*}
\sign({\theta_{ij}})\cdot \E_x X_{ij}\geq 2q\Big(|\theta_{ij}|\cdot \tfrac1{4} e^{-10 d \beta}e^{-Ld}- Ld\Big) 
\end{align*}
\item[(ii)] If $\{i,j\}\notin E$, then for any $x\in \{-1,+1\}^p$,
$$|\E_x X_{ij}|\leq 2q
Ld\,.$$
\end{enumerate}
\end{lemma}
\vspace{2mm}
\begin{proof} 
To begin, conditioning on $D_{ij}$ gives
\begin{align}
\E_x X_{ij}
=
\E_x(X_{ij}|D_{ij}) \P_x(D_{ij}) + \E_x(X_{ij}|D_{ij}^c) \P_x(D_{ij}^c)\label{e:Xexpansion}
\,.
\end{align}
In both cases (i) and (ii) we have
\begin{align}
\label{e:temp1}
|\E_x(X_{ij}|D_{ij}^c) \P_x(D_{ij}^c)| &\stackrel{(a)}{\leq}
 2\P_x(C_{ij}|D_{ij}^c)\P_x(D_{ij}^c)\nonumber
 \\
&\stackrel{(b)}{\leq}
2\P_x(A_{ij}|D_{ij}^c)P_x(D_{ij}^c) \nonumber
\\&\stackrel{(c)}{=}  2\P_x(A_{ij})P_x(D_{ij}^c) \nonumber
 \\& \stackrel{(d)}{\leq} 2q( 1-e^{-Ld})
 \\&\stackrel{(e)}{\leq} 2q Ld
\,.
\end{align}
Inequality (a) is by the crude estimate $|(-1)^{\ind\{\sig_j(L/3)=+1\}}(\sigma_i(L/3) - \sigma_i(L))|\leq 2$, (b) follows from the containment $C_{ij}\subseteq A_{ij}$, (c) is by independence of $A_{ij}$ and $D_{ij}$,  (d) is obtained by plugging in \eqref{e:A1prob} and \eqref{e:Bprob}, and (e) follows from the inequality $e^{-t}\geq 1-t$.

We first prove case (ii). 
If edge $\{i,j\}$ is not in the graph, then flipping only spin $\sig_j$ does not change the conditional distribution of spin $\sig_i$, assuming the neighbors of $i$ remain unchanged, and it follows from \eqref{e:edgeEffectEst} that 
$$\E_x (X_{ij}|D_{ij}) = 0\,.$$
Plugging \eqref{e:temp1} into \eqref{e:Xexpansion} proves case (ii).

We now turn to case (i). Suppose $\{i,j\}\in E$. Eq.~\eqref{e:Xexpansion} implies
\begin{align}
\sign(\theta_{ij}) \cdot \E_x(X_{ij})
\nonumber
&\geq\sign(\theta_{ij}) \cdot\E_x(X_{ij}|D_{ij}) \P(D_{ij})\\&\quad - |\E_x(X_{ij}|D_{ij}^c) \P(D_{ij}^c)|\label{e:Xexpansion2}
\,.
\end{align}
The second term has already been bounded in \eqref{e:temp1}. 
We estimate the first term on the right-hand side of~\eqref{e:Xexpansion2}: 
\begin{align*}
&\sign(\theta_{ij})\cdot \E_x(X_{ij}|D_{ij})\P(D_{ij})
\\&\stackrel{(a)}{=}
2\sign(\theta_{ij})\big(\pp(x_{\partial i\setminus \{j\}}) - \pmi(x_{\partial i\setminus \{j\}})\big)
 \P_x(C_{ij}|D_{ij})  \P(D_{ij}) 
\\&\stackrel{(b)}{\geq} 2\big(e^{4|\theta_{ij}|}-1\big)\cdot \tfrac1{16} e^{-10d\beta}\P(A_{ij})\P(D_{ij})
\\&\stackrel{(c)}{\geq} 2\cdot 4|\theta_{ij}|\tfrac1{16} e^{-10d\beta}q e^{-Ld}
\,.
\end{align*}
Here (a) 
uses~\eqref{e:edgeEffectEst}, (b) is by~\eqref{e:squeeze}
and because the reasoning from \eqref{e:AprobEst} applies also conditioned on $D_{ij}$ and using the fact that $A_{ij}$ and $D_{ij}$ are independent, and (c) follows from the inequality $e^x\geq 1+x$, the definition $q=\P(A_{ij})$, and \eqref{e:Bprob}.
This proves part (i). 
\end{proof}


We will use the following Bernstein-type submartingale concentration inequality, which can be found for example as an implication of Theorem~27 in~\cite{chung2006concentration}.
\begin{lemma}\label{l:concentration}
Let $Z_1,\dots, Z_n$ be a submartingale adapted to the filtration $(\mathcal{F}_k)_{k\geq 0}$ with
$|Z_k - Z_{k-1}|\leq c$ almost surely and $\text{Var} ( Z_k|\mathcal{F}_{k-1})\leq \sigma^2$. 
Then for all $N\geq 0$ and real $t$, 
$$
\P(Z_N - Z_0\leq -t)\leq \exp \left(  - \frac{t^2}{2N\sigma^2 + ct/3}\right).
$$
\end{lemma}
\vspace{2mm}

We now prove Theorem~\ref{t:Structure}.

\begin{proof} 
Recall that $q = [(1-e^{-L/3})e^{-L/3})]^3$. 
Suppose that $\{i,j\}\in E$. Let $\rho$ denote the lower bound quantity in case (i) of Lemma~\ref{l:goodEstimate}.
The inequality $e^{-t}\geq 1-t$ implies that 
\begin{align*}
\rho 
&= 2q\Big(|\theta_{ij}|\cdot \tfrac1{4} e^{-10 d \beta}e^{-Ld}- Ld\Big)
\\&\geq 2q\Big( \frac\al{4} e^{-10 d \beta}e^{-Ld}- Ld\Big)
\\&\geq 
2q(4Lde^{-Ld} - Ld)\geq 4qLd
\,.
\end{align*}
Here we used the bound $L\leq \beta e^{-10\beta}/16d\leq 1/160d$ so $e^{-Ld}\geq e^{-1/160}\geq 3/4$.

The sequence $Z_k = \sum_{\ell = 1}^{k}\sign(\theta_{ij})X_{ij}^{(\ell)} - k\rho$, $k\geq 1$, is a submartingale adapted to the filtration $(\mathcal{F}_k)_{k\geq 1}=(\sigt{0}{kL})_{k\geq 1}$, since by Lemma~\ref{l:goodEstimate},
$$
\E ( \sign(\theta_{ij})X_{ij}^{(k)}| \sigma^{((k-1)L)}) \geq \min_x \{\sign(\theta_{ij})\E_x X_{ij}\}\geq \rho\,.
$$

Let $\overbar{X}_{ij} = \frac1{\kmax}\sum_{k=1}^{\kmax} X_{ij}^{(k)}$.
Define $\sigma^2$ as in Lemma~\ref{l:concentration}, and note that $|Z_k-Z_{k-1}|\leq 2+ \rho\leq 3$. Recalling the choice $\tau = 3Ldq$, by Lemma~\ref{l:concentration} 
\begin{align*}
\P(\sign(\theta_{ij})\overbar{X}_{ij} 
< \tau)
 &=\P(Z_{\kmax}
 < \kmax (\tau - \rho))
 \\&\leq
 \P(Z_{\kmax}
 < -\kmax Ldq)
\\&\leq  \exp\bigg(-\frac{\kmax(Ldq)^2}{2 \sigma^2 + Ldq} \bigg)\,.
\end{align*}
It remains to bound $\text{Var} ( Z_k|\mathcal{F}_{k-1})$. For this, we observe that $\text{Var} ( Z_k|\mathcal{F}_{k-1})\leq 4 \cdot \P (X^{(k)}_{ij}\neq 0)$ (since $X^{(k)}_{ij}\in \{-2,0,2\}$). Now, $\P (X^{(k)}_{ij}\neq 0)\leq \P(C_{ij})\leq q$, so $\sigma^2\leq 4q$. 
We therefore obtain
$$
\P(\sign(\theta_{ij})\overbar{X}_{ij} 
< \tau) \leq \exp( - \kmax L^2 d^2 q/ 9)\,,
$$
where we used the crude bound $Ld\leq 1$. 

If $\kmax = 27(L^2d^2 q)\inv \log p$ then we can take a union bound over the at most $pd/2\leq p^2$ edges to see that with probability at least $1-\frac1p$ we have $E\subseteq \widehat E$. We can translate this value for $\kmax$ to the time $T$ stated in the theorem by taking $T$ larger than 
$$ 
\frac{64^2\cdot 27\cdot 6}{\alpha^2}e^{20d\beta} \log p =\frac{27\cdot 6}{L^2 d^2}\log p
\geq 
\frac{27}{L d^2 q}\log p
=L\kmax\,.
$$
The inequality used the estimate which holds for $L\leq 1/2$:
$
1/q\leq 3e^L L\inv\leq 6 L\inv\,.
$

Next, suppose that $\{i,j\}\notin E$. 
Lemma~\ref{l:goodEstimate} states that
$
|\E_x X_{ij}| \leq 2q
 ( 1-e^{-Ld}) \leq 2Ldq:=\rho'\,.
$
As before this implies that
$Z_k=\sum_{\ell=1}^{k}X_{ij}^{(\ell)} - k\rho'$, $k\geq 1$, is a supermartingale and $\widetilde{Z}_k=\sum_{\ell=1}^{k}X_{ij}^{(\ell)} + k\rho'$, $k\geq 1$, is a submartingale. 
Lemma~\ref{l:concentration} gives
\begin{align*}
\P(|\overbar{X}_{ij}| \geq \tau) 
&\leq \P(Z_{\kmax}\geq \kmax (\tau -\rho'))
\\ &\quad+  \P(\widetilde{Z}_{\kmax}\leq \kmax ( \rho'-\tau))
\\&\leq 2 \exp\bigg(-\frac{\kmax (Ldq)^2}{2\sigma^2 + Ldq } \bigg)\,.
\end{align*}
The same bound on $\sigma^2$ applies as before, and a union bound over at most ${p\choose 2}$ non-edges 
shows that the same $\kmax$ (and hence $T$) specified earlier suffices in order that $\widehat E\subseteq E$ with probability $1-\frac1p$.   
\end{proof}

\section{A lower bound on the observation time}
\label{s:lower}

Our lower bound derivation is a modification of the proof of Santhanam and Wainwright \cite{santhanam2012information} for the i.i.d. setting.
Their construction was based on cliques of size $d+1$, with a single edge removed. When the interaction is ferromagnetic (i.e., $\theta_{ij}\geq 0$), at low temperatures ($\alpha,\beta$ large enough) the removal of a single edge  is difficult to detect and leads to a lower bound.

We use a similar (but not identical) family of models as in \cite{santhanam2012information} to lower bound the observation time required. Start with a graph $G_0$ consisting of $\lfloor p/(d+1)\rfloor$ cliques of size $d+1$. Suppose that $d$ is odd, and fix a perfect matching on each of the cliques (each matching has cardinality $(d+1)/2$).  The vector of parameters $\theta^0$ corresponding to $G_0$ is obtained by setting $\theta_{ij}^0=\alpha$ for edges in the matchings, and $\theta_{ij}^0 = \beta$ for edges not in the matchings.

Now for each $\{u,v\}$ in a matching (where $\theta^0_{uv}=\alpha$) we form the graph $G_{uv}$ by removing the edge $\{u,v\}$ from $G_0$.
There are 
$$
M = \left\lfloor \frac p{d+1}\right \rfloor \left(\frac{d+1}{2}\right)\geq \frac{p}4
$$
graphs $G_{uv}$ with one edge removed. 

This construction is a refinement of the one in \cite{santhanam2012information}: their construction had all edge parameters equal to a single value $\beta$, and therefore did not fully capture the effect of some edges being dramatically weaker. 

\begin{theorem}[Sample complexity lower bound]\label{t:lower}
Suppose the minimax risk is $R_{p,d,n}\leq 1/2$. Then  $T=n/p$ satisfies
$$
T\geq \frac{e^{2\beta d/3}}{32e^6\alpha} \log p\,.
$$
\end{theorem}
\vspace{3mm}

In the remainder of this section we prove Theorem~\ref{t:lower}. We use the following version of Fano's inequality, which can be found, for example, as Corollary~2.6 in \cite{tsybakov2009introduction}. It gives a lower bound on the error probability (minimax risk in our case) in terms of the KL-divergence between pairs of points in the parameter space, where KL-divergence between two distributions $P$ and $Q$ on a space $\XX$ is defined as 
$$
D(P\|Q) = \sum_{x\in \XX} P(x) \log \frac{P(x)}{Q(x)}\,.
$$

\begin{lemma}[Fano's inequality]\label{e:Fano}
Assume that $M\geq 2$ and that $\Theta$ contains elements $\theta_0, \theta_1, \dots, \theta_M$. Let $Q_{\theta_j}$ denote the probability law of the observation $X$ under model $\theta_j$. 
If 
\begin{equation}
\label{e:KLbound}
	\frac1{M+1}\sum_{j=1}^M D(Q_{\theta_j}\|Q_{\theta_0})\leq \gamma\log M
\end{equation}  
for $0< \gamma < 1/8$, then the minimax risk for the zero-one loss is bounded as
$$
p_e\geq \frac{\log{(M+1)} - 1}{\log M} - \gamma\,.
$$
\end{lemma}

\subsection{Bound on KL divergence}
In this section we upper bound the KL divergence between the models parameterized by $\theta^0$ and any $\theta^{uv}$ (by symmetry of the construction this is the same for every $\theta^{uv}$). It suffices to consider the projection (i.e., marginal) onto the size $d+1$ clique containing $u$ and $v$, since the KL divergence between these projections is equal to the entire KL divergence. We therefore abuse notation slightly and write $\P_{\theta^0}$ and $\P_{\theta^{uv}}$ for the Gibbs distributions after projecting onto the relevant clique. Similarly, using $\theta$ as a placeholder for either $\theta^0$ or $\theta^{uv}$,
we let
$Q_{\theta}$ represent the distribution of the observation $X$, which now consists of samples $\sigma^{(1)},\dots,\sigma^{(n)}\in \{-1,+1\}^{d+1}$ as well as node update indices $I^{(1)},\dots,I^{(n)}\in [p]$. (We only project the Gibbs measure to the clique, keeping node update indices over the entire original graph.) The initial configuration $\sigma^{(1)}$ is drawn according to the stationary measure $\P _{\theta}$ for each model $Q_\theta$. Concretely, with $\theta$ representing either $\theta^0$ or $\theta^{uv}$,
\begin{align}
&Q_\theta (\sigma^{(1)},\dots,\sigma^{(n)}, I^{(1)},\dots,I^{(n)})\nonumber
 \\&\quad = \frac1{p^n} \cdot \P_{\theta}(\sigma^{(1)}) \prod_{l=2}^n \P _{\theta} (\sigma^{(l)}|\sigma^{(l-1)}, I^{(l)})\,.
 \label{e:Q}
\end{align}
Here the factor $1/p^n$ is due to updated node indices being uniformly random at each step. Implicit in the notation is the understanding that $\P _{\theta} (\sigma^{(l)}|\sigma^{(l-1)}, I^{(l)})=0$ if $\sigma^{(l-1)}$ and $\sigma^{(l)}$ differ in any spin other than $I^{(l)}$.

We have the following bound for each of the KL divergence terms in~\eqref{e:KLbound} (from which Theorem~\ref{t:lower} follows immediately from Lemma~\ref{e:Fano}).

\begin{lemma}\label{l:KLestimate}
For each model $Q_{\theta^{uv}}$ on graph $G_{uv}$,
$$
D(Q_{\theta^{uv}}\|Q_{\theta^0})\leq 4\al + \frac{n}p 18 \alpha d e^d e^{-2\beta d/3}\,.
$$
\end{lemma}\vspace{2mm}

\newcommand{\tuv}{\theta^{uv}}
\newcommand{\tz}{{\theta^{0}}}

\begin{proof}
Using~\eqref{e:Q} we write 
\begin{align}
D(Q_{\theta^{uv}}\| Q_{\theta^0})
&=\E_{X\sim Q_{\theta^{uv}}} \log \frac{Q_{\theta^{uv}}(X)}{Q _{\tz}(X)}\nonumber\\
&\!:=C_1 + \sum_{l=2}^n C_l\,,
\end{align}
where
\begin{align}
C_1 = \E_{\sig\sim \P_{\theta^{uv}}}  
 \log \frac{\P_{\theta^{uv}}(\sig)}{\P _{\tz}(\sig) }\label{e:C1}
\end{align}
and for $l\geq 2$
\begin{align}
C_l=\E_{\sig^{(l)},\sig^{(l-1)}\sim Q_{\theta^{uv}}}  
 \log \frac{\P_{\theta^{uv}}(\sig^{(l)}|\sig^{(l-1)},I^{(l)})}{\P _{\tz}(\sig^{(l)}|\sig^{(l-1)},I^{(l)} ) }\,.
 \label{e:KLexpansion}
\end{align}

Note that from any configuration $\sigma^{(l-1)}$, an update to node $k$ other than $u$ or $v$ has ratio of conditional probabilities equal to one (since the neighborhood of $k$ is the same under both models),
so each term in~\eqref{e:KLexpansion} is nonzero only if one of the nodes $u$ or $v$ is updated. This introduces a factor $2/p$ for the probability of selecting $u$ or $v$ to update, and by symmetry of the construction we can condition on $u$ updating. Thus,
\begin{align}
&C_l\nonumber
\\&=
\frac2p\E_{\sigma^{(l)}\sigma^{(l-1)}\sim Q_{\theta^{uv}}} \left[\log \frac{\P_{\theta^{uv}}(\sig_u^{(l)}|\sig^{(l-1)},I^{(l)})}{\P _{\tz}(\sig_u^{(l)}|\sig^{(l-1)},I^{(l)})}\bigg| I^{(l)}=u\right] 
.
\label{e:KLtemp1}
\end{align}

When updating node $u$ we have by \eqref{e:Glauber}
\begin{align}
&\frac{\P_{\theta^{uv}}(\sigma^{(l)}_u=+1|\sigma^{(l-1)},I^{(l)}=u)}{\P_{\theta^0}(\sigma_u^{(l)}=+1|\sigma^{(l-1)},I^{(l)}=u)}\nonumber
\\
&= \frac{1+ \exp(-2\alpha \sigma_v^{(l-1)}-2\beta\sum_{j\notin \{u,v\}} \sigma_j^{(l-1)})}{1+ \exp(-2\beta\sum_{j\notin \{u,v\}} \sigma_j^{(l-1)})}\label{e:condProbTemp}
\\& = \frac{\exp(2\beta\sum_{j\notin \{u,v\}} \sigma_j^{(l-1)}) + \exp(-2\alpha\sigma_v^{(l-1)})}{\exp(2\beta\sum_{j\notin \{u,v\}}\label{e:condProb} \sigma_j^{(l-1)})+1}\\&\leq e^{2\alpha}\,.\label{e:KL2alpha}
\end{align}
The summations indexed by ${j\notin \{u,v\}}$ are over nodes in the size $d+1$ clique under consideration.
The last inequality follows by observing that the largest value is achieved in \eqref{e:condProb} when $\sigma_v^{(l-1)} = -1$ and $\sum_{j\notin \{u,v\}} \sigma_j^{(l-1)} \to -\infty$. By symmetry 
the same bound holds for the ratio of conditional probabilities of $\sigma_u=-1$. 

 Equation~\eqref{e:KL2alpha} shows that the log-likelihood ratio is always at most $2\alpha$.
However, it is \emph{typically} roughly $e^{-c d \beta}$, where $c>0$ is a constant, because the effective magnetic field $\beta \sum_{j\notin \{u,v\}} \sigma_j$ typically has magnitude on the order~$\beta d$, as shown in Lemma~\ref{l:largeMag} later in this section.
Consider the event $U_l=\{ \sum \sigma_i^{(l-1)} \geq d/3+2\}$.
Applying the inequality $e^{2z}-1\leq 7z$ for $0\leq z\leq 1$ 
to~\eqref{e:condProb} gives
\begin{align}
&\frac{\P_{\theta^{uv}}(\sigma^{(l)}_u=+1|U_l,I^{(l)}=u)}{\P_{\theta^0}(\sigma_u^{(l)}=+1|U_l,I^{(l)}=u)} \nonumber
\\&\leq 1+  \frac{e^{2\alpha}-1}{1+\exp(2\beta d/3)}
	\nonumber
\\&\leq 1+7\alpha \exp e^{-2\beta d /3 }\,,\label{e:ratioBoundSmall}
\end{align}
and also from \eqref{e:Glauber}
\begin{equation}\label{e:ProbBoundSmall}
\P_{\theta^{uv}}(\sigma_u^{(l)}=-1|U_l,I^{(l)}=u) \leq e ^{-2\beta  d/3 }\,.
\end{equation}

We now bound each term $C_l$ in~\eqref{e:KLtemp1}. 
Let $\overbar{U}_l$ denote the event that $\big| \sum_{i} \sigma^{(l-1)}\big| \geq d/3+2$. Conditioning on~$\overbar {U}_l$ gives 
\begin{align}
C_l&=\E_{ Q_{\theta^{uv}}} \log \frac{\P_{\theta^{uv}}(\sig_u^{(l)}|\sig^{(l-1)},I^{(l)}=u)}{\P _{\tz}(\sig_u^{(l)}|\sig^{(l-1)},I^{(l)}=u)}\nonumber
\\&=
\E_{ Q_{\theta^{uv}}} \left[\log \frac{\P_{\theta^{uv}}(\sig_u^{(l)}|\sig^{(l-1)},I^{(l)}=u)}{\P _{\tz}(\sig_u^{(l)}|\sig^{(l-1)},I^{(l)}=u)}\bigg| \overbar U_l\right] \P_{\theta^{uv}}(\overbar U_l)\nonumber
\\&\quad + 
\E_{ Q_{\theta^{uv}}} \left[\log \frac{\P_{\theta^{uv}}(\sig_u^{(l)}|\sig^{(l-1)},I^{(l)}=u)}{\P _{\tz}(\sig_u^{(l)}|\sig^{(l-1)},I^{(l)}=u)}\bigg| \overbar U_l^c\right] \P_{\theta^{uv}}(\overbar U_l^c)\nonumber
\\ &= 
\E_{ Q_{\theta^{uv}}} \left[\log \frac{\P_{\theta^{uv}}(\sig_u^{(l)}|\sig^{(l-1)},I^{(l)}=u)}{\P _{\tz}(\sig_u^{(l)}|\sig^{(l-1)},I^{(l)}=u)}\bigg|  U_l\right] \P_{\theta^{uv}}(\overbar U_l)\nonumber
\\&\quad + 
\E_{ Q_{\theta^{uv}}} \left[\log \frac{\P_{\theta^{uv}}(\sig_u^{(l)}|\sig^{(l-1)},I^{(l)}=u)}{\P _{\tz}(\sig_u^{(l)}|\sig^{(l-1)},I^{(l)}=u)}\bigg| \overbar U_l^c\right] \P_{\theta^{uv}}(\overbar U_l^c)\label{e:Uconditioning}
\end{align}
The only change in the last equality is replacing $\overbar U_l$ by $U_l$ in the first conditional expectation, which is justified by symmetry of the model to flipping all the spins. 
Using \eqref{e:KL2alpha},\eqref{e:ratioBoundSmall}, \eqref{e:ProbBoundSmall}, $\log(1+x)\leq x$, and $\P_{\theta^{uv}}(\overbar U_l)\leq 1$, the first term in \eqref{e:Uconditioning} is bounded by
$$
2\al e^{-2\beta  d/3 } +7\alpha e^{-2\beta  d/ 3} \leq 9\alpha e^{-2\beta  d/ 3}\,.
$$
Using~\eqref{e:KL2alpha} and
Lemma~\ref{l:largeMag} below,
the second term in \eqref{e:Uconditioning} is bounded by 
$$
2\alpha\P_{\theta^{uv}}(\overbar U_l^c) \leq 2\alpha d(3e)^{\frac d3+1} \exp(-\beta d(d-3)/3)\,.
$$
Combining the last two displays gives
\begin{align*}
C_l &\leq 9\alpha e^{-2\beta d/ 3}+2\alpha d(3e)^{\frac d3+1} \exp(-\beta d(d-3)/3)
\\&\leq  9 \alpha d e^d e^{-2\beta d/3 }
\end{align*}
and adding this quantity $n$ times and multiplying by the factor $2/p$ in \eqref{e:KLtemp1}\,, 
we get
$$
D(Q_{\theta^{uv}}\| Q_{\theta^0})\leq C_1 + \frac{n}p 18 \alpha d e^d e^{-2\beta d/3 }\,.
$$

Now, to bound $C_1$, it suffices to bound ${\P_{\theta^{uv}}(\sig)}/{\P _{\tz}(\sig) }$. Let $g_{uv}(\sig) = Z_{uv}\P_{\theta^{uv}}(\sig)$ and $g_{0}(\sig) = Z_{0}\P_{\theta^{0}}(\sig)$, where 
$Z_{uv}=\sum_{\sig}g_{uv}(\sig)$ and $Z_{0}=\sum_{\sig}g_{0}(\sig)$ are the partition functions for the two models. An argument similar to~\eqref{e:KL2alpha} shows that $e^{-2\al} g_{uv}(\sig)
\leq g_0(\sig)\leq e^{2\al} g_{uv}(\sig)$ for any~$\sig$, hence 
$$
C_1\leq \log \frac{\P_{\theta^{uv}}(\sig)}{\P _{\tz}(\sig) }=\log \frac{Z_0\cdot  g_{uv}(\sig)}{Z_{uv}\cdot g _{0}(\sig) }\leq 4\al\,.
$$
Plugging this quantity into the previous displayed equation completes the proof. 
\end{proof}

\begin{lemma} \label{l:largeMag}
The magnetization $\sum_i \sigma_i$ satisfies 
$$
\P_{{\theta^{uv}}} (|\textstyle{\sum}_i{\sigma_i}| \leq d/3+1) \leq d(3e)^{\frac d3+1} \exp(-\beta d(d-3)/3)\,.$$  
\end{lemma}
\vspace{3mm}

\begin{proof}
	Note that $\P_{\theta^{uv}}$ is the stationary measure for the Glauber dynamics governing $Q_{\theta^{uv}}$, so the marginal distribution of each $\sig^{(l)}$ in the sample $X\sim Q_{\theta^{uv}}$ is $\P_{\theta^{uv}}$.
	
We first lower bound ${\P_{\theta^{uv}} (|\sum_i{\sigma_i}|> d/3+1)}$ by the probability of the all $+1$ or all $-1$ configuration,
\begin{align*}
&{\P_{\tuv} \Big(\Big|\sum_i{\sigma_i}\Big|> \frac d3+1\Big)} \\&\geq \frac2Z \exp\bigg(\beta \frac{(d-1)(d+1)}{2}+ \alpha \frac{d-1}{2}\bigg)
\\&\geq 
\frac2Z \exp\bigg(\beta \cdot \frac{d^2 -1}{2}\bigg)
\,.
\end{align*}
Next, by supposing all edges in the clique have parameter $\beta$ we get the upper bound
\begin{align*}
{\P_{\tuv}\Big(\big|\sum_i{\sigma_i}\big|\leq \frac d3+1\Big)} &\leq \frac {2d}Z {d\choose \frac d3+1} \exp(\beta d^2/9 + \beta d/2)\\&\leq (3e)^{\frac d3+1}\frac {2d}Z \exp(\beta d^2/9 + \beta d/2)\,,
\end{align*}
where the second inequality follows from
$
{n\choose k}\leq \left(\frac{n\cdot e}{k} \right)^k
$ and $(3e)^{d/3}\leq e^d$. 
Taking the ratio of the last two displayed quantities gives the desired inequality.
\end{proof}

\section{Discussion}
The main message of this paper is that observing dynamics over time is quite natural in many settings, and that access to such observations leads to a simple algorithm for estimating the graph underlying an Ising model. We expect that similar results can be derived (with suitable modifications) for samples generated from local Markov chains other than the Glauber dynamics, and for non-binary pairwise graphical models. Several other generalizations are plausible; for instance, it would be interesting to consider the situation where one only observes samples intermittently.

\section*{Acknowledgments}
We are grateful to Mina Karzand, Kuang Xu, and Luis Voloch for helpful comments on a draft of the paper, and to Vijay Subramanian for an interesting discussion on coordination games. 
This work was supported in part by NSF grants CMMI-1335155 and CNS-1161964, and by Army Research Office MURI Award W911NF-11-1-0036.

\bibliographystyle{ieeetr}
\bibliography{spinGlass_BIB}

\end{document}